\documentclass[letterpaper]{article} 
\usepackage{aaai2026}  
\usepackage{times}  
\usepackage{helvet}  
\usepackage{courier}  
\usepackage[hyphens]{url}  
\usepackage{graphicx} 
\usepackage{natbib}  
\usepackage{caption} 
\usepackage{algorithm}
\usepackage{algorithmic}
\usepackage{svg}
\usepackage{booktabs}
\usepackage{multirow}
\usepackage{array}
\usepackage{comment}
\usepackage{amsmath}
\usepackage{amsfonts}
\usepackage{subcaption}
\usepackage{tabularx}
\usepackage{float}
\usepackage{dblfloatfix}
\usepackage{adjustbox}
\usepackage[table]{xcolor}
\usepackage{diagbox}

\usepackage{newfloat}
\usepackage{listings}
\DeclareCaptionStyle{ruled}{labelfont=normalfont,labelsep=colon,strut=off} 
\floatstyle{ruled}
\newfloat{listing}{tb}{lst}{}
\floatname{listing}{Listing}

\title{Condensed Data Expansion Using Model Inversion for Knowledge Distillation}
\author{
    Kuluhan Binici\textsuperscript{\rm 1},
    Shivam Aggarwal\textsuperscript{\rm 2},
    Cihan Acar\textsuperscript{\rm 3},
    Nam Trung Pham\textsuperscript{\rm 4},
    Karianto Leman\textsuperscript{\rm 3},
    Gim Hee Lee\textsuperscript{\rm 2},
    Tulika Mitra\textsuperscript{\rm 2}
}
\affiliations{
    \textsuperscript{\rm 1}SAP
    \\
    \textsuperscript{\rm 2}National University of Singapore
    \\
    \textsuperscript{\rm 3}Institute for Infocomm Research, A*STAR
    \\
    \textsuperscript{\rm 4}Intelexvision
    
}

\usepackage{bibentry}

\begin{document}

\maketitle

\begin{abstract}
Condensed datasets offer a compact representation of larger datasets, but training models directly on them or using them to enhance model performance through knowledge distillation (KD) can result in suboptimal outcomes due to limited information. To address this, we propose a method that expands condensed datasets using model inversion, a technique for generating synthetic data based on the impressions of a pre-trained model on its training data. This approach is particularly well-suited for KD scenarios, as the teacher model is already pre-trained and retains knowledge of the original training data. By creating synthetic data that complements the condensed samples, we enrich the training set and better approximate the underlying data distribution, leading to improvements in student model accuracy during knowledge distillation. Our method demonstrates significant gains in KD accuracy compared to using condensed datasets alone and outperforms standard model inversion-based KD methods by up to  $11.4\%$  across various datasets and model architectures. Importantly, it remains effective even when using as few as one condensed sample per class, and can also enhance performance in few-shot scenarios where only limited real data samples are available.
\end{abstract}

\section{Introduction}
Condensed datasets \cite{zhao2021dataset} have emerged as a promising approach for compactly representing large datasets, enabling efficient model training with reduced memory and computational costs. These datasets consist of synthetic samples optimized to capture the information content of much larger datasets. They provide certain privacy benefits, as studied in \cite{dong2022privacy} and can be produced with modest memory and time resources through recent methods \cite{zhou2022dataset,zhao2023dataset,feng2023embarrassingly}. These qualities render condensed samples suitable for scenarios in which the privacy considerations prohibit exposure of individual training samples or the large memory size aggravating their relocation. However, the utility of condensed datasets at small scale can be limited in various learning paradigms \cite{yu2023dataset} such as standard supervised learning or knowledge distillation (KD) \cite{hinton2015distilling}. This limited utility stems from the reduced information captured in these compact representations, hindering the ability of models to effectively learn from them. In KD, this limitation is particularly relevant as a student model learns from the guidance of a pre-trained teacher model, whose ability to transfer knowledge is itself constrained by the limited information present in the condensed dataset.

\begin{figure}[!t]
  \centering
    \includegraphics[width=0.90\linewidth]{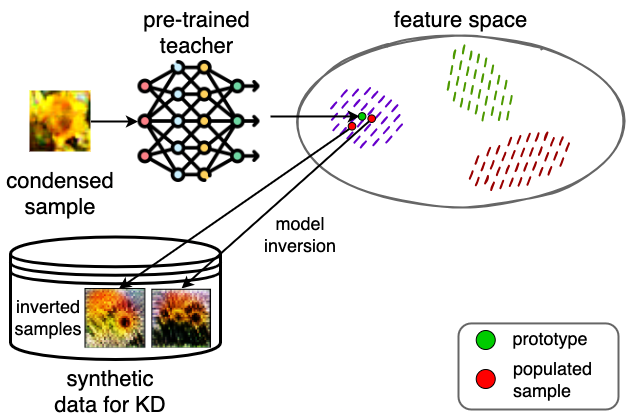} 
   \caption{ Illustration of our motivation for using condensed samples as prototypes for synthetic data. }
   \label{fig:motivation}
\end{figure}

In this work, we address the limited utility of condensed datasets, particularly in KD, by expanding them using model inversion (MI) \cite{lopes2017data,yu2023data,liu2024small}. Our goal is to enhance the limited information captured by these compact representations and better approximate the underlying training data. MI is a technique that leverages a pre-trained model as a discriminator to generate synthetic samples resembling real ones. This requirement of a pre-trained model is naturally satisfied in KD, as it inherently involves a pre-trained teacher model. MI operates by training a generative model to produce data points that follow the learned distribution of the teacher's representations \cite{liu2021data}. Since the true data distribution is unknown without access to the original dataset, the generative model is optimized based on certain inductive biases \cite{zhao2018bias,goyal2022inductive} about the training data. One such bias is the assumption that the teacher classifies real samples with high confidence, resulting in near one-hot prediction vectors \cite{chen2019data,yin2020dreaming}.

A trivial approach for expanding the condensed data through MI is simply combining it with the synthetic samples obtained by MI. However, our preliminary experiments suggest that this does not improve accuracy, which is likely caused by the domain gap between the two sample sets \cite{hennicke2024mind,bai2024bridging}. With this observation, in this paper we propose using the condensed samples as prototypes that represent the real data distribution and changing the MI process to generate samples that are more aligned with the characteristics of the condensed data. By prioritizing the generation of synthetic samples that resemble these prototypes, our method can bridge the domain gap and enhance the performance of the student model trained on the expanded data. Another reason contributing to such performance improvement is that the inductive bias used by MI to model the real data distribution is improved with the knowledge of the prototypes. 

Specifically, our method utilizes a small set of condensed samples to query the teacher model and extract more realistic impressions related to the target dataset. These samples are fed to the teacher model for estimating the per-class feature distributions of the target data. Then, the model inversion process is conditioned to produce synthetic samples following a similar feature distribution as the condensed ones. This is achieved by configuring a feature discriminator \cite{li2020adversarial} that competes against the sample generator to distinguish condensed samples from the ones generated by model inversion. By doing so, the generator is forced to produce synthetic samples with semantic similarity to the condensed ones, thus reducing the risk of a domain gap and improving accuracy. 
{\em One of the major advantages of our approach is its versatility, as it can be applied on top of any model inversion method to improve the accuracy of the student.} 
This is evident from the experimental evaluation, where we record accuracy improvements of up to $11.44\%$ compared to different state-of-the-art KD baselines across multiple model pairs and datasets. The effectiveness of our method is more pronounced for teacher-student pairs with little structural similarity. In addition, remarkably, even using as few as one condensed sample per class results in a noticeable accuracy improvement.

Moreover, our method is also applicable to scenarios where a limited amount of real samples from the training set are accessible, such as in few-shot learning scenarios \cite{song2023comprehensive,sauer2022knowledge}. Experimental results exhibit the advantage of using synthetic data samples generated with guidance from the real samples against pure few-shot KD methods.

\section{Related Work}
\subsection{Dataset Condensation}
Dataset condensation was introduced by \cite{zhao2021dataset} to reduce the training time required for large-scale datasets. It optimizes small batches of synthetic samples to carry almost equal information content as real batches of much larger size. The resulting samples are typically not visually realistic and can better protect data privacy than communicating real samples \cite{dong2022privacy}. To quantitatively assess the privacy benefits, \cite{zhou2022dataset} exercised membership inference attacks (MIA) using condensed samples and showed they yield only around $0.52$ attack AUC, which is almost the same value as random guessing, i.e. $0.5$.
Therefore, in cases where samples from a real training set cannot be communicated due to privacy concerns, these condensed samples can be utilized to guide KD's synthetic data generation process. Moreover, the time cost of dataset condensation has significantly decreased due to recent advancements in the area \cite{zhao2023dataset,feng2023embarrassingly}. As such, \cite{zhou2022dataset} can generate a condensed dataset of size 10 samples-per-class from ImageNet-200 in less than an hour with 2 GB memory utilization on a single Nvidia Quadro RTX 6000. This allows condensed versions of datasets at various scales (ranging from MNIST to ImageNet) to be conveniently produced and released to public. \cite{hemultisize} further reduces the storage requirement of dataset condensation by compressing multiple condensation processes into a single one.

\begin{figure*}[!th]
  \centering
    \includegraphics[width=0.90\linewidth]{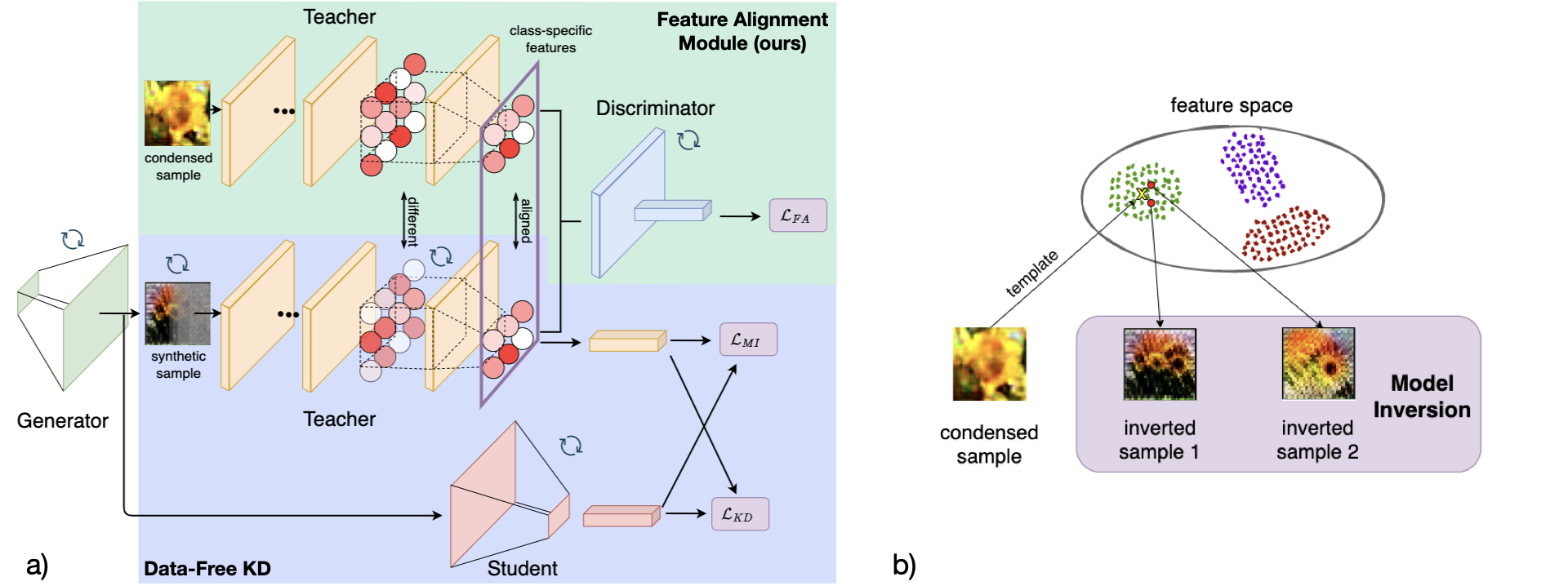} 
   \caption{(a) Overview of our condensed-samples guided GMI (generative model inversion) framework. The discriminator is optimized to distinguish real and fake features, while the generator tries to prevent it from doing so by aligning them. Generator, discriminator, and student models are trained in alternate steps.  (b) Illustration of our motivation for using condensed samples as templates for synthetic data.}
   \label{fig:framework}
\end{figure*}

\subsection{Knowledge Distillation (KD)}
KD \cite{hinton2015distilling} trains a compact ``student'' neural network model to approximate the decision space of a more complex one called the ``teacher''. The inclusion of the soft guidance supplied by the teacher enriches the limited information the student receives from the one-hot encoded class labels. As a result, the student can achieve better performance than supervised training alone can provide. While most commonly, the logit scores or the softmax probabilities of the teacher are considered for regularization \cite{romero2014fitnets}, activation maps or attention scores can also be used \cite{zagoruyko2016paying}.
\subsection{Model Inversion}
If the training dataset is entirely inaccessible, conventional KD methods can not operate. 
The textit {Model Inversion} (MI) technique is developed to infer data samples that the teacher model had observed during training and use them for distillation. Some early works consider the confidence of teachers' predictions as supervision for sample generation \cite{chen2019data}. 
Some others generate samples that maximize the information gain to the student \cite{micaelli2019zero}. Following these, DeepInversion \cite{yin2020dreaming} proposed taking advantage of the batch normalization statistics gathered while training the teacher model. CMI \cite{fang2021contrastive} improved on this method by diversifying sample synthesis with the help of contrastive learning. Fast-Datafree \cite{fang2022up} proposed a technique to reduce the significant amount of time that model inversion takes. PRE-DFKD \cite{binici2022robust} introduced a method to eliminate the trade-off between the large memory footprint and the robustness of the process. Recent works have focused on further addressing scalability and efficacy issues \cite{yu2023data,liu2024small}.
\subsection{Few-Shot KD}
In few-shot KD, only a small subset of real training samples are accessible. To avoid over-fitting, available methods typically reduce the number of parameters required to train the student network.
\cite{bai2020few} feeds activation of teacher and student networks to the layers of one another for cross-correction. FSKD \cite{li2020few} obtains the student architecture from the teacher itself via pruning, freezes it, and only learns 1x1 convolutions added after each layer. NetGraft \cite{shen2021progressive} performs distill

\section{Condensed Data Expansion Using MI}
Model inversion techniques \cite{yin2020dreaming,chen2019data,fang2021contrastive} typically use variants of the following loss function to guide the synthetic data generation. 
\begin{equation}
\resizebox{.9\hsize}{!}{$
\mathcal{L}_{MI} = \mathbb{E}_{x, y_{ps}}\left[y_{ps}log(\hat y_{T})\right] 
-\mathbb{E}_{x}\left[D_{KL}(\hat{y_{T}}||\hat{y_{S}})\right]
$}
\label{eq1_1}
\end{equation}
The first term targets to maximize the softmax score ($\hat{y}_{(T)}$) (confidence) that the synthetic samples $\hat{x}$ receive from the teacher $T$ for the pseudo-classes ($y_{ps} \sim u(0,c)$) assigned to them. Generally, $\hat{x}$ is obtained using a generative model parameterized by $\theta_g$ ($x\sim p_{\theta_{g}}(x|z)$). 
The last term encourages the synthesis of samples that provide high information gain to the student. As this objective is strictly guided by the knowledge that teacher embodies, prototype information cannot be incorporated to provide additional guidance. 
\subsection{Feature alignment mechanism:}
To incorporate the condensed data prototypes form the training distribution in the model inversion objective, we add the constraint of aligning the feature distribution of generated synthetic samples with that of the condensed ones. This constraint is enforced through the inclusion of a feature discriminator in our model inversion framework. As illustrated in Figure \ref{fig:framework}, first, the generator outputs synthetic samples. Then, the teacher model encodes both these synthetic batches and condensed samples into feature representations. Later, discriminator classifies these features as condensed or synthetic samples. This classification results in a \textit{feature alignment} loss that quantifies the gap between synthetic and condensed feature distributions. This can be viewed as a minimax game in which the feature discriminator competes against the generator for distinguishing real features from synthetic ones. In equilibrium, the generator will be able to provide samples that can yield similar features as the real ones to trick the discriminator. Essentially, the condensed samples serve as prototypes, guiding the generation of new samples that reflect the training distribution.

The final optimization objective for the generator is constructed by combining the feature alignment loss with the objective of any base model inversion method $\mathcal{L}_{MI}$ as shown in Equation \ref{eq2}.
\vspace{1pt}
\begin{equation}
\begin{split}
&\min\mathcal{L}_{G} = \mathcal{L}_{MI} + \mathcal{L}_{FA} \\
&\resizebox{.9\hsize}{!}{$
\max\mathcal{L}_{D} = \mathbb{E}_{\hat{x}}\left[\log D(\phi_l(\hat{x}))\right] + \mathbb{E}_{\hat{x}}\left[1 - \log D(\phi_l(\hat{x}))\right]
$}
\label{eq2}
\end{split}
\end{equation}
\par \noindent
Where $\mathcal{L}_{FA} = \mathbb{E}_{\hat{x}}\left[1 - \log D(\phi_l(\hat{x}))\right]$ stands for feature alignment loss. As the dimension of feature vectors is high with respect to the limited availability of condensed samples, our method is prone to over-fitting. Therefore we use a simple discriminator architecture introduced by \cite{li2020adversarial} that contains very few parameters. To further address the risk of over-fitting, we perform differentiable data augmentation \cite{zhao2020differentiable} on both the synthetic images output by the generator and the condensed samples before feeding them to the discriminator. 
\par
In deciding on the layer index at which the feature alignment will be employed, we considered the type of image features encoded by different parts of the teacher model. Typically, for image inputs, early layers of neural networks encode structural patterns that are commonly shared across natural images (e.g., edges). In contrast, the image features occurring at the later layers contain semantical information. As our objective is to produce diverse views of objects from the same semantical classes as the condensed samples, we considered features at late layers, specifically the penultimate layer, of the teacher as alignment targets. 
\par
Additionally, we posit that simply aligning the cumulative distribution of synthetic features from all classes with the real feature distribution is not ideal. Rather, we provide class-specific alignment using a conditional discriminator \cite{mirza2014conditional}. The contrast between these two alternatives can be seen in the figure given in the appendix. This further changes our discriminator objective to the following.
\begin{equation}
\begin{split}
\max\mathcal{L}_{D} &= \mathbb{E}_{(x,y)}\left[\log D(\phi_l(x), y)\right] \\
&+ \mathbb{E}_{(\hat{x}, y_{ps})}\left[1 - \log D(\phi_l(\hat{x}),  y_{ps})\right]
\label{eq1_2}
\end{split}
\end{equation}
Here, the discriminator not only predicts if a feature is associated with a condensed or synthetic sample but also determines the class it belongs to. To enforce this, we present three different types of inputs to the discriminator. First, we construct ``real'' inputs by pairing real features with their labels. Later, we use the teacher to assign labels to the synthetic features and obtain ``fake'' inputs. Lastly, to prevent the discriminator from neglecting the class information, we construct additional ``fake'' inputs by pairing the same real features with the wrong class labels. Formally, our real ($\mathcal{R}$) and fake ($\mathcal{F}$) sets can be defined as,
\begin{equation}
\begin{split}
\mathcal{R} &= \{(x,y)| (x, y) \in \mathcal{X}\} \\
\mathcal{F} &= \{(\hat{x}, y_{ps})\} \cup \{(x,c)| (x, y) \in \mathcal{X}, c \neq y\}
\label{eq1_3}
\end{split}
\end{equation}
\subsection{Combining condensed and synthetic samples:}
\label{sec:comb_cond_syn}
After establishing our feature-alignment strategy to improve model inversion with the available data samples, we discuss how we can join condensed and generated synthetic samples for KD. 
Some alternatives included pre/post-training the student with condensed samples with respect to model inversion. 
However, as these methods can cause the student to be biased towards one data type, we avoided them. 
Instead, we expanded the condensed dataset by adding the iteratively refined synthetic samples (through model inversion) and trained the student with randomly sampled batches from such union.
Our distillation objective involves minimizing the distance between the predictions of the teacher and the student models, which can be summarized as,
\begin{equation}
\resizebox{.9\hsize}{!}{
$\theta^{*}_{S} := \arg\min\limits_{\theta_S}\mathbb{E}_{\hat{x}}\left[D_{KL}(\hat{y}_{S}||\hat{y}_{T})\right] + \mathbb{E}_{x}\left[D_{KL}(y_{S}||y_{T})\right]
$}
\label{eq:eqKD}
\end{equation}
In Equation \ref{eq:eqKD}, $\hat{x}$ and $x$ denote synthetic samples and condensed samples respectively. The exact procedure we follow in generating synthetic samples and distilling the student is summarised in Algorithm \ref{alg:GFSKD}. First, we initialize our synthetic dataset $\mathcal{X}$ with the available condensed samples. Later at each epoch, we generate a new synthetic batch via our condensed sample-guided model inversion and add it to $\mathcal{X}$. Later, within the same epoch, we randomly draw a data batch from $\mathcal{X}$ and use it to transfer knowledge from the teacher to the student.
\setlength{\intextsep}{10pt}
\setlength{\textfloatsep}{10pt}
\begin{algorithm}[!h]
\footnotesize
\caption{Knowledge Distillation}
\begin{algorithmic}
\STATE \text{\textbf{INPUT:}} generator $G$, discriminator$D$, teacher $T$, student $S$ parameterized by $\theta_{S}$, condensed data $\mathcal{X}$
\STATE \text{\textbf{OUTPUT:}} trained student $\theta^{*}_{S}$.
\newline
\FOR{number of epochs}
    \STATE $\hat{x}_{new} \gets invert\_model(T, S, G, D, \mathcal{X})$
    \STATE  $\mathcal{X} \gets	\mathcal{X} \cup \hat{x}_{new}$
    \STATE $(x, y) \sim \mathcal{X}$
    \STATE $\mathcal{L}_{KD} \gets \sum\limits_{\mathcal{X}}D_{KL}(\hat{y}_{S}||\hat{y}_{T})$
    \STATE $optimizer.step(backward(\mathcal{L}_{KD}),\theta_{S})$
\ENDFOR
\end{algorithmic}
\label{alg:GFSKD}
\end{algorithm}
\vspace{-5pt}
\section{Experimental Evaluation}
To assess the effectiveness of our method, we incorporate condensed data guidance to three state-of-the-art model-inversion methods and record the improvement in KD performance. 
These methods are Fast, CMI, and PRE-DFKD. 
Moreover, we also experiment with applying out method to expand limited real data and observe the advantage against few-shot KD methods. 
For this comparison, we selected NetGraft and FSKD as baselines. All the results we report on the performance of our baseline methods are either directly taken from the papers or obtained by running the official implementations based on the hyper-parameter configurations shared in the papers or GitHub pages. To standardize the evaluation, we use fixed random seeds borrowed from the official implementations of the baselines. 

\begin{table*}[!t]
\centering
\resizebox{1\textwidth}{!}{
\begin{tabular}{lcccccccccc}
\hline
\multicolumn{1}{l|}{Dataset}                                                               & \multicolumn{4}{c|}{CIFAR-10}                                                          & \multicolumn{4}{c|}{CIFAR-100}                                                         & \multicolumn{2}{c}{ImageNet-200} \\ \hline
\multicolumn{1}{l|}{Teacher}                                                               & ResNet-34      & ResNet-34      & WRN-40-2       & \multicolumn{1}{c|}{WRN-40-2}       & ResNet-34      & ResNet-34      & WRN-40-2       & \multicolumn{1}{c|}{WRN-40-2}       & ResNet-34       & ResNet-34      \\
\multicolumn{1}{l|}{Student}                                                               & ResNet-18      & MBNet-v2       & WRN-40-2       & \multicolumn{1}{c|}{MBNet-v2}       & ResNet-18      & MBNet-v2       & WRN-40-2       & \multicolumn{1}{c|}{MBNet-v2}       & ResNet-18       & MBNet-v2       \\ \hline
\multicolumn{1}{l|}{Teacher acc.}                                                          & 95.70          & 95.70          & 94.87          & \multicolumn{1}{c|}{94.87}          & 78.05          & 78.05          & 75.83          & \multicolumn{1}{c|}{75.83}          & 71.20           & 71.20          \\ \hline
\multicolumn{11}{c}{\cellcolor[HTML]{C0C0C0}Training student (S) with labeled data}                                                                                                                                                                                                                             \\
\multicolumn{1}{l|}{\begin{tabular}[c]{@{}l@{}}Train w/ full  real data\end{tabular}}    & 95.20          & 93.79          & 94.87          & \multicolumn{1}{c|}{93.79}          & 77.10          & 72.80          & 75.83          & \multicolumn{1}{c|}{72.80}          & 64.90           & 55.06          \\
\multicolumn{1}{l|}{\begin{tabular}[c]{@{}l@{}}Train w/ cond. samples (CS)\end{tabular}} & 34.66          & 30.13          & 38.92          & \multicolumn{1}{c|}{30.13}          & 16.54          & 10.69          & 15.55          & \multicolumn{1}{c|}{10.69}          & 3.50            & 5.60           \\ \hline
\multicolumn{11}{c}{\cellcolor[HTML]{C0C0C0}Distilling student (S) with synthetic data}                                                                                                                                                                                                                         \\
\multicolumn{1}{l|}{Fast}                                                                  & 92.62          & 86.12          & 92.82          & \multicolumn{1}{c|}{85.06}          & 69.76          & 54.62          & 65.05          & \multicolumn{1}{c|}{48.21}          & 42.99           & 35.31          \\
\multicolumn{1}{l|}{Fast + CS}                                                             & 92.72          & 86.37          & 92.84          & \multicolumn{1}{c|}{85.69}          & 69.96          & 56.57          & 65.51          & \multicolumn{1}{c|}{49.42}          & 45.43           & 40.68          \\
\multicolumn{1}{l|}{Fast* (ours w/ CS)}                                                          & \textbf{94.33} & \textbf{88.05} & \textbf{94.64} & \multicolumn{1}{c|}{\textbf{89.24}} & \textbf{72.09} & \textbf{63.29} & \textbf{70.96} & \multicolumn{1}{c|}{\textbf{60.86}} & \textbf{48.14}  & \textbf{43.08} \\ \hline
\multicolumn{1}{l|}{CMI}                                                                   & 94.84          & 87.5           & 92.83          & \multicolumn{1}{c|}{86.53}          & 77.04          & 61.9           & 68.96          & \multicolumn{1}{c|}{59.04}          & 44.11           & 35.55          \\
\multicolumn{1}{l|}{CMI + CS}                                                              & 94.89          & 88.06          & 92.94          & \multicolumn{1}{c|}{86.77}          & 77.04          & 62.54          & 69.10          & \multicolumn{1}{c|}{59.62}          & 47.07           & 40.67          \\
\multicolumn{1}{l|}{CMI* (ours w/ CS)}                                                           & \textbf{94.97} & \textbf{89.63} & \textbf{94.21} & \multicolumn{1}{c|}{\textbf{90.21}} & \textbf{77.07} & \textbf{70.21} & \textbf{72.42} & \multicolumn{1}{c|}{\textbf{68.05}} & \textbf{48.98}  & \textbf{45.83} \\ \hline
\end{tabular}
}
\caption{Student accuracies (\%) obtained by expanding condensed data using Fast and CMI. Condensed datasets with 50 spc from CIFAR10, 10 spc from CIFAR100, and 10 spc from ImageNet-200 were used.
}
\label{table:fast_comparison}
\end{table*}
\begin{figure*}[!b]
  \centering
  \begin{subfigure}[b]{0.45\textwidth}
    \centering
    \includegraphics[width=0.75\textwidth]{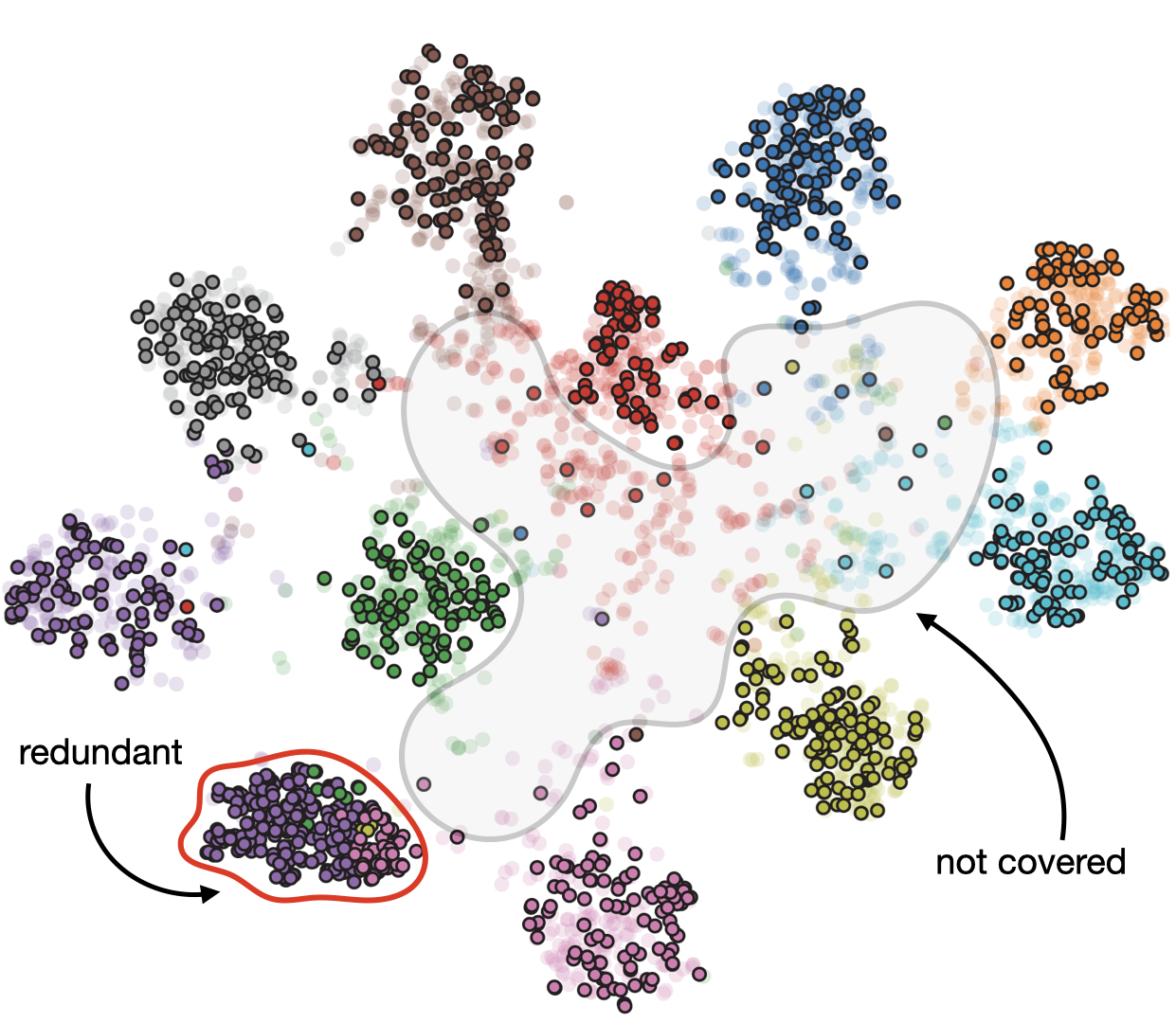}
    \caption{Synthetic feature distribution by data-free model inversion}
  \end{subfigure}%
  \hspace{15pt}
  \begin{subfigure}[b]{0.45\textwidth}
    \centering
    \includegraphics[width=0.75\textwidth]{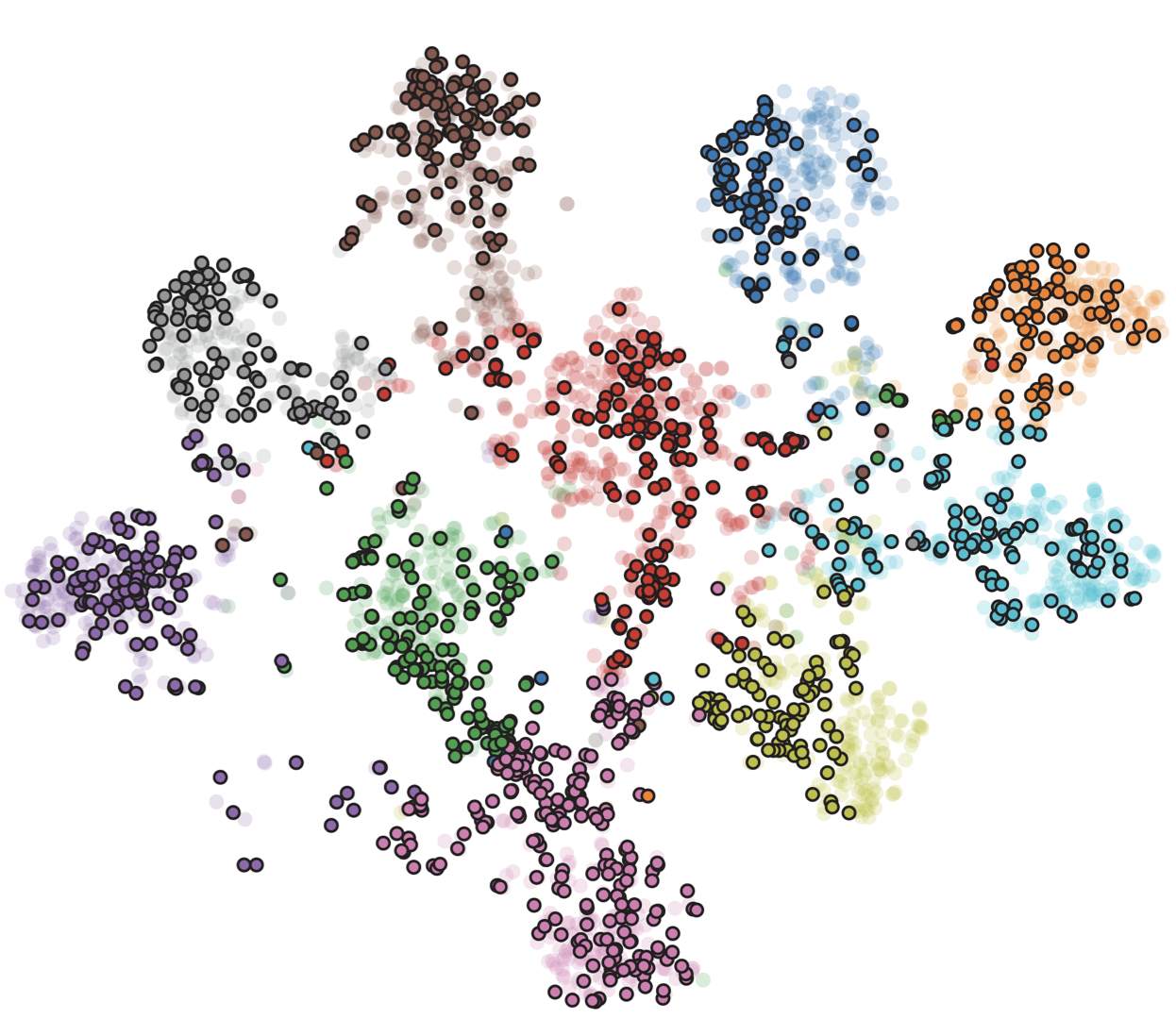}
    \caption{Condensed data-guided synthetic feature distribution (ours)}
  \end{subfigure}
  \caption{2D visualisation of feature vectors. Faded and bold markers denote feature space projections of real samples from the CIFAR-10 dataset and synthetic samples, respectively. Condensed data-guided synthetic samples exhibit better alignment with the real data distribution.}
  \label{fig:tsne}
\end{figure*}

\subsubsection{Datasets}
We use three image classification datasets, which are CIFAR-10/100 \cite{krizhevsky2009learning}, and ImageNet-200 \cite{deng2009imagenet}. We conducted our experiments using condensed samples generated by three different methods, including those provided by \cite{zhao2021siamese}, \cite{cazenavette2022dataset} and \cite{zhao2023dataset}. While the results reported throughout our experiments primarily utilize the condensed samples from \cite{zhao2021siamese}, we also include an ablation study to compare the effectiveness of each condensation method.

\subsubsection{Implementation details}
We used the same generator architectures as shared in the official implementations of the MI methods that we couple our method with (Fast, CMI, and PRE-DKD). Further details on generator and discriminator architectures as well as how we couple our method with individual MI methods can be found in the appendix. 
\subsection{Impact of expanded condensed data on KD}
Tables \ref{table:fast_comparison} and \ref{tab:PRE-DFKD-CS} show the student accuracies upon coupling our approach with Fast, CMI, and PRE-DFKD. The results achieved by this coupling are indicated by the asterisk symbol with the annotation ``*(ours w/ CS)''.
We also configure naive baselines where we combine these samples with the synthetic datasets generated by model inversion methods. These are denoted with ``+ CS'' notations. 
Further, the row ``Train w/ full real data'' represents the accuracy of full-scale training of students on the target dataset and constitutes the upper bound. ``Train w/ cond. samples'' reflects the accuracy achieved by only using condensed samples for student training, which is the lower bound. 
In all experiments, we use condensed datasets of 50 samples per class (spc) (total 500 samples) and 10 spc (total 1000 samples) for CIFAR-10 and CIFAR-100, respectively. For ImageNet-200, we use 10 spc (total 2000 samples). These correspond to 1\%, 2\%, and 2\% of the total samples in their respective datasets.

First, we note that the performance of Fast method notably diminishes for pairs with low structural similarity (e.g. ResNet-34 \cite{he2016deep} \& MobileNet-v2 \cite{howard2017mobilenets}). The improvement was especially significant for WRN-40-2 \cite{zagoruyko2016wide} \& MobileNet-v2 pairs reaching up to $11.44\%$ on CIFAR-100. CMI also has a considerable performance gap with respect to the upper bound for heterogeneous model pairs.
Our method again achieves consistent advantage, with substantial accuracy improvements reaching up to $8.43\%$ (WRN-40-2 \& MobileNet-v2). 
Since PRE-DFKD achieves almost the same student accuracy as the upper limit for homogeneous pairs (e.g. ResNet-34 \& ResNet-18), we only experiment with heterogeneous ones (e.g. ResNet-34 \& MobileNet-v2), where there is still room for improvement. The results in Table \ref{tab:PRE-DFKD-CS} shows that our method effectively improves acuracy also for this baseline method.
\par
In all experiments, the simple combination of condensed samples and the synthetic samples from model inversion (``+ CS'') neither mitigated this issue nor caused any substantial performance improvement in most cases. On the other hand, our condensed sample-guided model inversion (``*'') consistently increased student accuracy across different datasets and teacher-student pairs, which ensures that the benefit of our approach is not simply due to exposing the student to more samples during training.

\begin{figure*}[!b]
  \centering
    \includegraphics[width=0.33\textwidth]{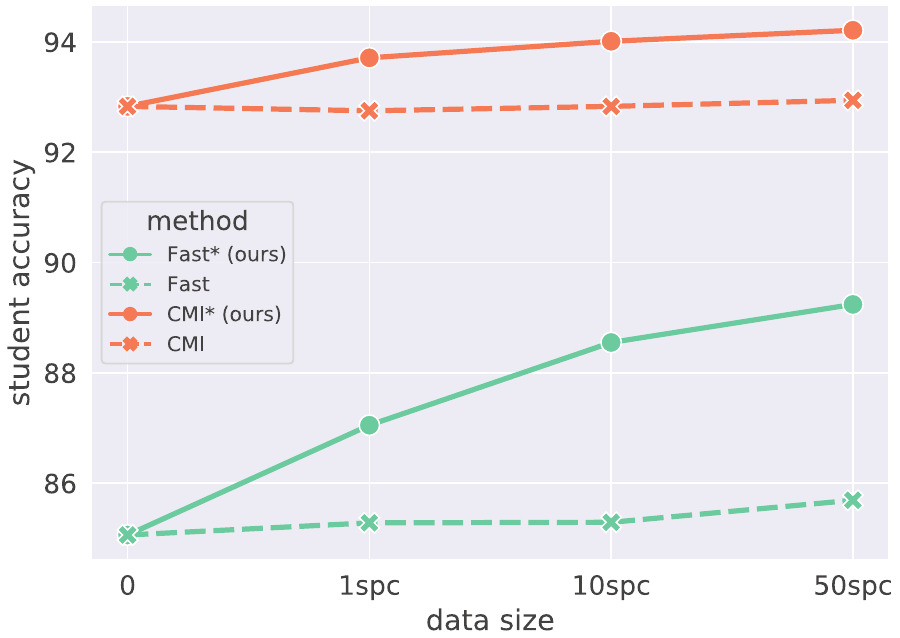}
    \hspace{25pt}
    \includegraphics[width=0.33\textwidth]{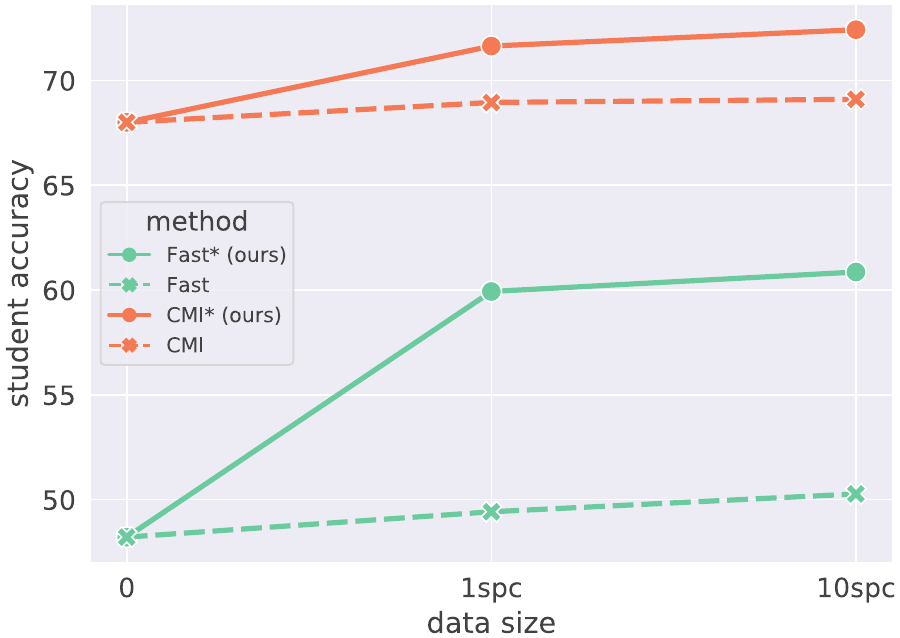}
   \caption{Students distilled by only using samples from model inversion (left), and using expanded data generated by our method (right) for different condensed dataset set sizes. Condensed datasets with 1, 10, and 50 spc from CIFAR10, and 1 and 10 spc from CIFAR100, were used.}
   \label{fig:acc_scale}
\end{figure*}
\vspace{-3pt}
\subsubsection{Visual results}
\begin{figure}[!t]
  \centering
    \includegraphics[width=0.85\columnwidth]{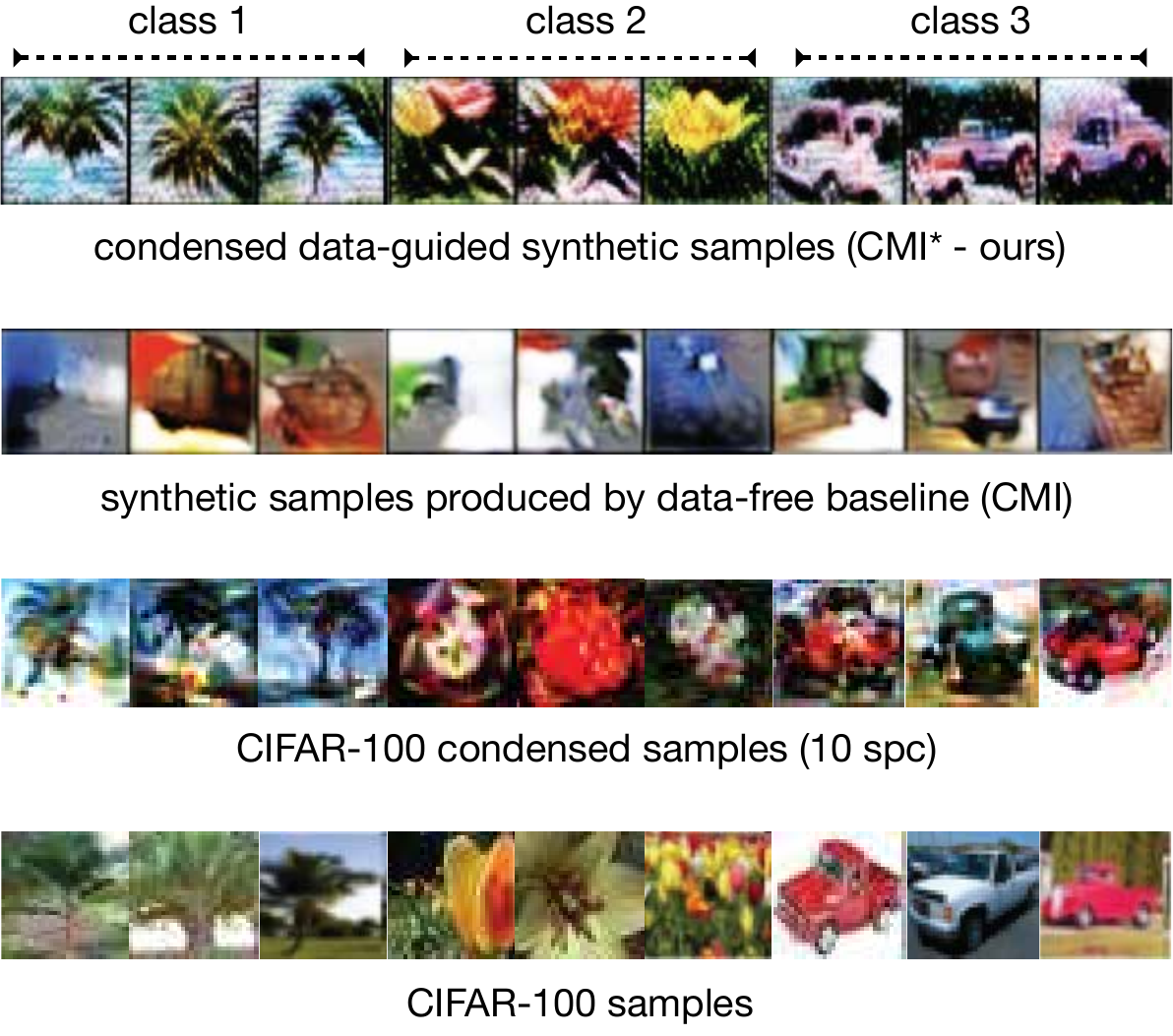}
   \caption{First two rows contain synthetic CIFAR-100 samples obtained with and w/o condensed data-guided model inversion. The last two show condensed and real samples.
   }
   \label{fig:vis_samples}
\end{figure}

We examine the impact of our method on the visual quality of the generated samples. Figure \ref{fig:vis_samples} contains synthetic CIFAR100 images obtained by CMI and CMI*. We note that CMI* samples are significantly more realistic and exhibit common class-distinctive patterns across images from the same categories, which is not observed in CMI. This strengthens the claim that feature alignment effectively conditions the synthetic data to contain realistic semantics that is consistent among samples from the same classes. 
Additionally, this conditioning does not compromise the diversity of the synthetic set, as demonstrated by the varied object views and scales in CMI*. We also note that neither the condensed samples nor the condensed sample-guided generated synthetic samples exist in the real dataset. Thus, although the generated synthetic samples are visually realistic, they do not reveal any individual training samples. Moreover, to visually observe the effect of our method on the distribution of generated synthetic samples, we projected the high-dimensional feature space on a 2D plane using the t-SNE algorithm \cite{van2008visualizing}. On the resulting plane, we compare the feature distributions observed for CMI and CMI*. We use the WRN-40-2 teacher trained on the CIFAR-10 dataset as the feature extractor. The visualizations are displayed in Figure \ref{fig:tsne} where faded data points mark the features of real samples, and the bold ones represent the synthetic ones. Each sample is marked with the color associated with its class label. Synthetic samples generated by CMI are mostly clustered together and only partially correspond with real samples. Also, some feature clusters have almost no correspondence with real features (circled in red), meaning  the associated synthetic samples might not be contributing to the knowledge transfer. In contrast, the synthetic feature distribution formed during CMI* shows better alignment with real data distribution. 
\paragraph{How does student accuracy scale with the condensed data size?} After establishing that our proposed method can boost the utility of condensed samples in KD, we analyze how the scale of the available data affects the improvement. For this, we consider condensed datasets of 3 different sizes (1 spc, 10 spc, 50 spc) for CIFAR-10 and 2 different sizes (1 spc, 10 spc) for CIFAR-100. These amounts correspond to $0.02\%$, $0.2\%$ and $1\%$ of the samples contained in CIFAR-10, and $0.2\%$ and $2\%$ for CIFAR-100. The plots in Figure \ref{fig:acc_scale} show that Fast and CMI baselines do not benefit from the mere inclusion of condensed samples in their synthetic distillation sets, irrespective of the number of samples available. However, when they are equipped with our feature alignment module, the student accuracies scale up with increasing data availability.
\paragraph{Does the condensation method affect the quality of model inversion?}
We study the impact of the dataset condensation used to produce the condensed samples on the effectiveness of our approach and display the results in Table \ref{table:condensed_ablation}.  
\begin{table}[!t]
\centering
\resizebox{\columnwidth}{!}{
\begin{tabular}{lcccccc}
\hline
\multicolumn{1}{l|}{Dataset}                                  & \multicolumn{6}{c}{CIFAR-10} \\ \hline
\multicolumn{1}{l|}{Teacher}                                  & \multicolumn{3}{c|}{ResNet-34}                          & \multicolumn{3}{c}{WRN-40-2}                          \\
\multicolumn{1}{l|}{Student}                                  & \multicolumn{3}{c|}{MobileNet-v2}                       & \multicolumn{3}{c}{MobileNet-v2}                      \\ \hline
\multicolumn{7}{c}{\cellcolor[HTML]{C0C0C0}Ablation Study on Different Dataset Condensation Methods}                                                                                                               \\ \hline
\multicolumn{1}{l|}{\diagbox[width=9em]{\textbf{KD Method}}{\textbf{CS Type}}} & \textbf{DSA} & \textbf{DM} & \multicolumn{1}{c|}{\textbf{MTT}} & \textbf{DSA} & \textbf{DM} & \textbf{MTT} \\ \hline
\multicolumn{1}{l|}{Fast + CS}                                & 86.37           & 86.54          & \multicolumn{1}{c|}{86.78}           & 85.69          & 80.01          & 80.38           \\ \hline
\multicolumn{1}{l|}{\textbf{Fast* (ours w/ CS)}}                & \textbf{88.05}  & \textbf{87.82} & \multicolumn{1}{c|}{\textbf{87.11}} & \textbf{89.24}  & \textbf{90.30} & \textbf{90.38} \\ \hline
\end{tabular}
}
\caption{Impact of different types of condensed samples (CS) on student accuracy (\%) for different dataset condensation strategies. Dataset contains 50 spc from CIFAR10.
}
\label{table:condensed_ablation}
\end{table}
DSA, DM and MTT refer to \cite{zhao2021siamese}, \cite{zhao2023dataset}, and \cite{cazenavette2022dataset} respectively. The results indicate that while the choice of condensation method can impact the final student accuracy, our method consistently improves performance across all types of condensed samples tested. This suggests that its effectiveness is not dependent on any single condensation approach.

\begin{table*}[!t]
\centering
\resizebox{0.82\textwidth}{!}{
\centering
\begin{tabular}{lcccccc}
\hline
\multicolumn{1}{l|}{Dataset}                        & \multicolumn{2}{c|}{CIFAR-10}                        & \multicolumn{2}{c|}{CIFAR-100}                       & \multicolumn{2}{c}{ImageNet-200}                  \\ \hline
\multicolumn{1}{l|}{Teacher}                        & ResNet-34      & \multicolumn{1}{c|}{WRN-40-2}       & ResNet-34      & \multicolumn{1}{c|}{WRN-40-2}       & ResNet-34      & ResNet-34                        \\
\multicolumn{1}{l|}{Student}                        & MobileNet-v2   & \multicolumn{1}{l|}{MobileNet-v2}   & MobileNet-v2   & \multicolumn{1}{l|}{MobileNet-v2}   & ResNet-18      & \multicolumn{1}{l}{MobileNet-v2} \\ \hline
\multicolumn{1}{l|}{Teacher acc.}                   & 95.70          & \multicolumn{1}{c|}{94.87}          & 78.05          & \multicolumn{1}{c|}{75.83}          & 71.20          & 71.20                            \\ \hline
\multicolumn{7}{c}{\cellcolor[HTML]{C0C0C0}Accuracy of student (S) trained with labeled data}                                                                                                                         \\
\multicolumn{1}{l|}{Train w/ full real data}        & 93.79          & \multicolumn{1}{c|}{93.79}          & 72.80          & \multicolumn{1}{c|}{72.80}          & 64.90          & 55.06                            \\
\multicolumn{1}{l|}{Train w/ cond. samples (CS)}    & 27.24          & \multicolumn{1}{c|}{27.24}          & 10.69          & \multicolumn{1}{c|}{10.69}          & 3.50           & 5.60                             \\
\multicolumn{1}{l|}{Train w/ few real samples (RS)} & 27.55          & \multicolumn{1}{c|}{27.55}          & 4.66           & \multicolumn{1}{c|}{4.66}           & 1.42           & 1.30                             \\ \hline
\multicolumn{7}{c}{\cellcolor[HTML]{C0C0C0}Accuracy of student (S) trained with synthetic data}                                                                                                                       \\
\multicolumn{1}{l|}{PRE-DFKD}                       & 83.12          & \multicolumn{1}{c|}{83.12}          & 66.56          & \multicolumn{1}{c|}{61.83}          & 54.20          & 47.26                            \\
\multicolumn{1}{l|}{PRE-DFKD* (ours w/ CS)}          & \textbf{86.89} & \multicolumn{1}{c|}{\textbf{88.34}} & \textbf{67.39} & \multicolumn{1}{c|}{\textbf{63.41}} & \textbf{55.22} & \textbf{49.73}                   \\
\multicolumn{1}{l|}{PRE-DFKD* (ours w/ RS)}          & \textbf{86.77} & \multicolumn{1}{c|}{\textbf{87.47}} & \textbf{71.63} & \multicolumn{1}{c|}{\textbf{64.37}} & \textbf{54.95} & \textbf{49.68}                   \\ \hline
\end{tabular}}
\caption{Student accuracies (\%) obtained by utilizing the expanded condensed and real datasets set using PRE-DFKD (CS and RS). Condensed and real sample sets contain 50 spc from CIFAR10, 10 spc from CIFAR100, and 10 spc from ImageNet-200.}
\label{tab:PRE-DFKD-CS}
\end{table*}
\begin{table*}[!h]
\centering
\small
\begin{tabularx}{\textwidth}{@{}X@{}}
\begin{minipage}[!t]{0.63\textwidth}
    \centering
    \resizebox{\textwidth}{!}{
    \begin{tabular}{lcccc}
    \hline
    Dataset      & \multicolumn{2}{c}{CIFAR-10} & \multicolumn{2}{c}{CIFAR100} \\ \hline
    Teacher      & VGG-11 & VGG-11 & VGG-11 & VGG-11 \\
    Student      & VGG-11 (50\% pruned) & VGG-11 (75\% pruned) & VGG-11 (25\% pruned) & VGG-11 (50\% pruned) \\ \hline
    Teacher acc. & 92.25  & 92.25  & 71.23  & 71.23  \\ \hline
    \multicolumn{5}{c}{\cellcolor[HTML]{C0C0C0}{\color[HTML]{333333} Few-shot distillation accuracy}} \\
    FSKD         & 78.69  & 36.00  & 54.72  & 24.73  \\
    PRE-DFKD*    & \textbf{83.26} & \textbf{68.48} & \textbf{63.81} & \textbf{61.21} \\ \hline
    \end{tabular}
    }
    \subcaption{Student accuracy comparison with FSKD baseline. Real sample sets containing 50 spc from CIFAR10, and 10 spc from CIFAR100 were used.}
    \label{tab:FSKD}
\end{minipage}
\hfill 
\begin{minipage}[!t]{0.35\textwidth}
    \centering
    \resizebox{0.70\textwidth}{!}{
    \begin{tabular}{lcc}
    \hline
    Dataset      & CIFAR-10 & CIFAR100 \\ \hline
    Teacher      & VGG-16   & VGG-16   \\
    Student      & ResNet-18 & ResNet-18 \\ \hline
    Teacher acc. & 94.16    & 74.00    \\ \hline
    \multicolumn{3}{c}{\cellcolor[HTML]{C0C0C0}Few-shot distillation accuracy} \\
    NetGraft     & 73.69    & 55.51    \\
    PRE-DFKD*    & \textbf{88.38} & \textbf{68.54} \\ \hline
    \end{tabular}
    }
    \subcaption{Comparison with few-shot NetGraft baseline. 1 spc from both CIFAR10 / 100 were used.}
    \label{tab:NetGraft}
\end{minipage}
\end{tabularx}
\caption{Comparison with different few-shot baselines using few real samples.}
\vspace{-10pt}
\end{table*}
\subsubsection{Ablation study}
To demonstrate the individual contributions of the constituents of method to the final performance, we provide an ablation study in Table \ref{table:condensed_ablation}.
\begin{table}[!h]
\vspace{5pt}
\centering
\resizebox{0.40\textwidth}{!}{
\begin{tabular}{cc}
\hline
Method                                                 & Student acc. \\ \hline
\multicolumn{1}{c|}{CMI}                               & 92.83        \\
\multicolumn{1}{c|}{+ CS}                              & 92.94        \\
\multicolumn{1}{c|}{+ CS-guided MI}                    & 93.98        \\
\multicolumn{1}{c|}{+ class-specific alignment (CMI*)} & \textbf{94.21}        \\ \hline
\end{tabular}
}
\caption{MobileNet-v2 student accuracy on CIFAR10 for various configurations of the components used in our method. Dataset with 50 spc from CIFAR10 was used.}
\label{tab:ablation}
\end{table}
This involves student accuracies upon enhancing a base model inversion method (CMI) in 3 different settings. (1) utilizing condensed samples only by appending them to the distillation set (``+ CS''), (2) using them also to regularize the feature distribution of synthetic data (``+ CS-guided MI'') (3) the final version of our condensed sample-guided model inversion (CMI*). As seen in Table \ref{tab:ablation} ``+ CS-guided MI'' achieves advantage over simply including condensed samples in the distillation set (``+ CS''). Later, we include the final component, i.e., class-specific alignment, and record student accuracy (``+ class-specific alignment''). This caused only a marginal increase in performance but we retain it as it improves results without introducing any notable computational burden.
\vspace{-11pt}
\subsection{Impact of expanded real data on KD}
Our method is also inherently capable of expanding limited real samples from the target dataset in few-shot scenarios. To show this, we repeated the experiments in Table \ref{tab:PRE-DFKD-CS} by replacing the condensed samples with real ones. We assume the availability of the same amount of samples randomly drawn from the target datasets as we used in experiments with condensed data (50 spc for CIFAR10, 10 spc for CIFAR100 and 10 spc for ImageNet-200). 
The results are displayed in Table \ref{tab:PRE-DFKD-CS} 
 as PRE-DFKD* (ours w/ RS). Similar to condensed sample experiments, the improvements observed for heterogeneous pairs are again greater than the homogeneous ones, i.e. ResNet-34 \& ResNet-18. 
 The overall improvement yielded by our method upon utilizing few real samples is comparable that using condensed ones. 

\subsubsection{Comparison with few-shot methods}
We benchmarked the effectiveness of our method in few-shot KD against FSKD and Netgraft baselines using the same experiment setups reported in their papers. To ensure a fair evaluation of our work against these few-shot baselines, we selected identical teacher-student pairs to those in the original papers. We again use the same amount of samples from both datasets as in Table \ref{tab:PRE-DFKD-CS}. In our comparison with FSKD, we use VGG11 teachers and student models that are channel-pruned versions of the teacher at different rates. As for the comparison with Netgraft, we use VGG16 teachers and ResNet-18 students, with 1 spc from both datasets. From Table \ref{tab:FSKD}, it can be observed  FSKD is only effective for low pruning rates while performing poorly for higher rates. Similarly, the accuracies achieved by NetGraft were also much lower than the upper-bound as seen in Table \ref{tab:NetGraft}. This is expected as both baselines are restricted to the limited information made available by the few-shot sample set. As our method has a generative component, it does not have such restriction and therefore outperforms both few-shot baselines by a large margin.
\vspace{-5pt}
\section{Conclusion}
We address the  challenge of limited information in condensed datasets, which often hinders their effectiveness in KD. We propose a method that leverages  MI to expand these condensed datasets, generating synthetic data that enriches the compact representation and improves KD performance. Our approach utilizes condensed samples as prototypes to guide the MI process, ensuring that the generated synthetic data aligns closely with the underlying data distribution represented by the condensed set. 
Our experiments demonstrate the effectiveness of our method across various datasets, model architectures, and state-of-the-art MI techniques (Fast, CMI, PRE-DFKD). 
The results show consistent improvements in KD, compared to using condensed datasets alone or standard MI-based KD methods. 
Our approach also applies to few-shot learning, outperforming existing few-shot KD by effectively leveraging limited real data for synthetic data generation.

\bibliography{aaai2026}

\newpage
\setlength{\leftmargini}{20pt}
\makeatletter\def\@listi{\leftmargin\leftmargini \topsep .5em \parsep .5em \itemsep .5em}
\def\@listii{\leftmargin\leftmarginii \labelwidth\leftmarginii \advance\labelwidth-\labelsep \topsep .4em \parsep .4em \itemsep .4em}
\def\@listiii{\leftmargin\leftmarginiii \labelwidth\leftmarginiii \advance\labelwidth-\labelsep \topsep .4em \parsep .4em \itemsep .4em}\makeatother

\setcounter{secnumdepth}{0}
\renewcommand\thesubsection{\arabic{subsection}}
\renewcommand\labelenumi{\thesubsection.\arabic{enumi}}

\newcounter{checksubsection}
\newcounter{checkitem}[checksubsection]

\newcommand{\checksubsection}[1]{%
  \refstepcounter{checksubsection}%
  \paragraph{\arabic{checksubsection}. #1}%
  \setcounter{checkitem}{0}%
}

\newcommand{\checkitem}{%
  \refstepcounter{checkitem}%
  \item[\arabic{checksubsection}.\arabic{checkitem}.]%
}
\newcommand{\question}[2]{\normalcolor\checkitem #1 #2 \color{blue}}
\newcommand{\ifyespoints}[1]{\makebox[0pt][l]{\hspace{-15pt}\normalcolor #1}}

\end{document}